\documentclass[sigconf]{acmart}
\AtBeginDocument{%
  }

\setcopyright{acmlicensed}
\copyrightyear{2018}
\acmYear{2018}
\acmDOI{XXXXXXX.XXXXXXX}
\acmConference[Conference acronym 'XX]{Make sure to enter the correct
  conference title from your rights confirmation email}{June 03--05,
  2018}{Woodstock, NY}
\acmISBN{978-1-4503-XXXX-X/2018/06}

\usepackage{booktabs}     
\usepackage{makecell}  
\usepackage{tabularx}      
\usepackage{pifont}       
\usepackage{multirow}     
\usepackage{caption}      
\usepackage{adjustbox}    
\usepackage{tcolorbox}

\usepackage{ragged2e}     
\usepackage{CJKutf8}
\usepackage{array}     
\newcommand{\cmark}{\ding{51}}
\newcommand{\xmark}{\ding{55}}
\newcommand{\yes}{\ding{51}}   
\newcommand{\no}{\ding{55}}    
\newcommand{\partialyes}{$\triangle$} 
\usepackage{float}         
\newcommand{\lbl}[1]{\textbf{\textsf{\scriptsize #1}}}

\begin{document}

\title{From Words to Worlds: Benchmarking Cross-Cultural Cultural Understanding in Machine Translation}

\author{Bangju Han}
\authornote{Both authors contributed equally to this research.}
\authornotemark[2]
\authornotemark[3]

\email{hanbangju23@mails.ucas.ac.cn}
\orcid{1234-5678-9012}
\affiliation{%
  \institution{Xinjiang Technical Institute of Physics \& Chemistry, Chinese Academy of Sciences}
  \city{Urumqi}
  \country{China}
}

\author{Yingqi Wang}

\authornotemark[1]
\email{wangyingqi23@mails.ucas.ac.cn}
\authornote{University of Chinese Academy of Sciences, Beijing, China (second affiliation)}
\authornote{Xinjiang Laboratory of Minority Speech and Language Information Processing, Urumqi, China (third affiliation)
}
\affiliation{%
  \institution{Xinjiang Technical Institute of Physics \& Chemistry, Chinese Academy of Sciences}
  \city{Urumqi}
  \country{China}
}

\author{Qing Huang}
\authornotemark[2]
\authornotemark[3]
\email{huangqing24@mails.ucas.ac.cn}
\affiliation{%
  \institution{Xinjiang Technical Institute of Physics \& Chemistry, Chinese Academy of Sciences}
  \city{Urumqi}
  \country{China}
}

\author{Tiyuan Li}
\authornotemark[2]
\authornotemark[3]
\affiliation{%
  \institution{Xinjiang Technical Institute of Physics \& Chemistry, Chinese Academy of Sciences}
  \city{Urumqi}
  \country{China}
}

\author{Fengyi Yang}
\authornotemark[2]
\authornotemark[3]
\affiliation{%
  \institution{Xinjiang Technical Institute of Physics \& Chemistry, Chinese Academy of Sciences}
  \city{Urumqi}
  \country{China}
}
\author{Ahtamjan Ahmat}
\authornotemark[2]
\authornotemark[3]
\affiliation{%
  \institution{Xinjiang Technical Institute of Physics \& Chemistry, Chinese Academy of Sciences}
  \city{Urumqi}
  \country{China}
}
\author{Abibulla Atawulla}
\authornotemark[2]
\authornotemark[3]
\affiliation{%
  \institution{Xinjiang Technical Institute of Physics \& Chemistry, Chinese Academy of Sciences}
  \city{Urumqi}
  \country{China}
}
\author{Ran Bi}
\authornotemark[2]
\authornotemark[3]
\affiliation{%
  \institution{Xinjiang Technical Institute of Physics \& Chemistry, Chinese Academy of Sciences}
  \city{Urumqi}
  \country{China}
}

\author{Yating Yang}
\authornotemark[2]
\authornotemark[3]
\affiliation{%
  \institution{Xinjiang Technical Institute of Physics \& Chemistry, Chinese Academy of Sciences}
  \city{Urumqi}
  \country{China}
}
\author{Xi Zhou}
\authornotemark[2]
\authornotemark[3]
\authornote{Corresponding author}
\affiliation{%
  \institution{Xinjiang Technical Institute of Physics \& Chemistry, Chinese Academy of Sciences}
  \city{Urumqi}
  \country{China}
}

\renewcommand{\shortauthors}{Trovato et al.}

\begin{abstract}


Culture-expressions, such as idioms, slang, and culture-specific items (CSIs), are pervasive in natural language and encode meanings that go beyond literal linguistic form. Accurately translating such expressions remains challenging for machine translation systems. Despite this, existing benchmarks remain fragmented and do not provide a systematic framework for evaluating translation performance on culture-loaded expressions. To address this gap, we introduce CulT-Eval, a benchmark designed to evaluate how models handle different types of culturally grounded expressions. CulT-Eval comprises over 7,959 carefully curated instances spanning multiple types of culturally grounded expressions, with a comprehensive error taxonomy covering culturally grounded expressions. Through extensive evaluation of large language models and detailed analysis, we identify recurring and systematic failure modes that are not adequately captured by existing automatic metrics. Accordingly, we propose a complementary evaluation metric that targets culturally induced meaning deviations overlooked by standard MT metrics. The results indicate that current models struggle to preserve culturally grounded meaning and to capture the cultural and contextual nuances essential for accurate translation. Our benchmark and code are available at \url{https://anonymous.4open.science/r/CulT-Eval-E75D/}.


\end{abstract}



\received{20 February 2007}
\received[revised]{12 March 2009}
\received[accepted]{5 June 2009}


\maketitle
\begin{figure}[t]
    \centering
    \includegraphics[width=1\linewidth]{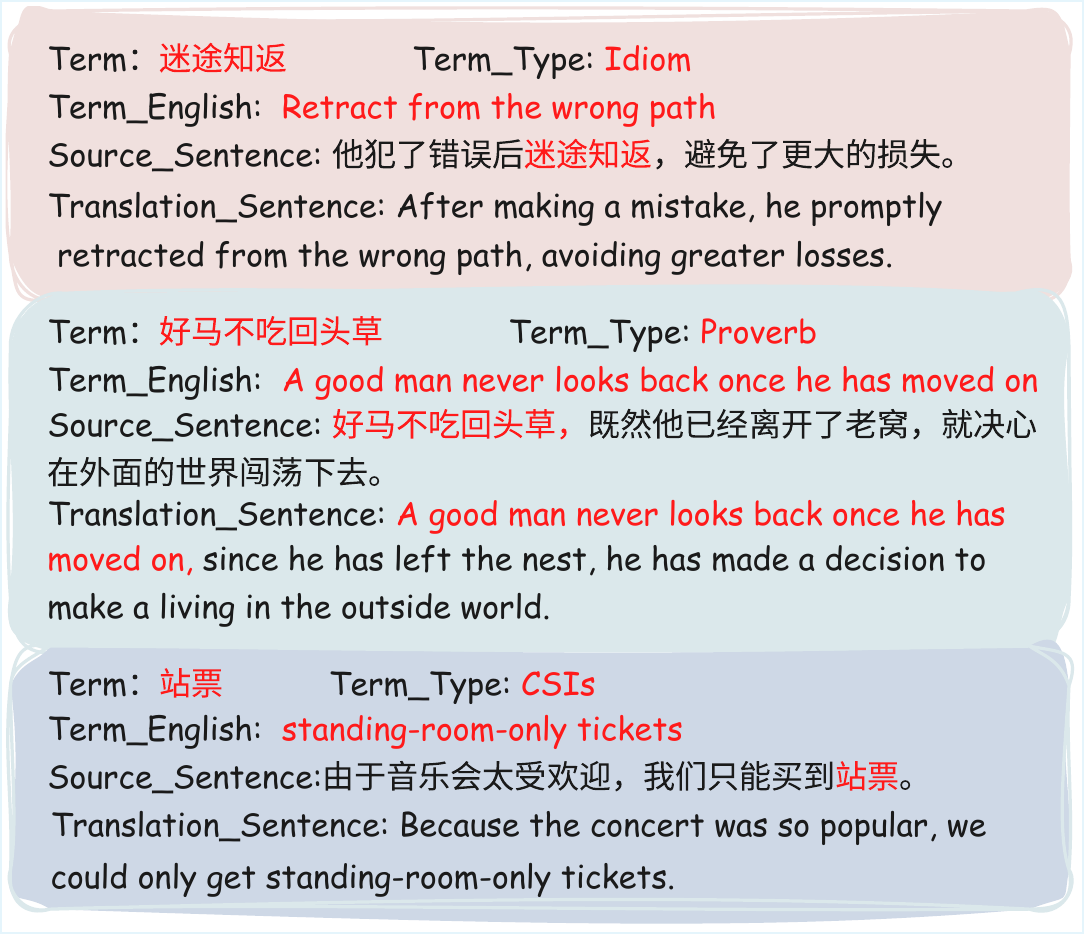}
    \caption{Representative CulT-Eval instances.}
    \label{fig:placeholder}
\end{figure}

\section{Introduction}
Rooted in shared cultural knowledge and social conventions, many expressions in natural language convey meanings that rely on implicit cultural knowledge rather than explicit linguistic forms. We refer to these expressions as culture-loaded expressions, such as idioms, slang, literary allusions, and culture-specific items (CSIs). Figure 1 presents representative instances across idioms, proverbs, and culture-specific items. Recent studies indicate that cultural grounding poses persistent challenges for modern large language models. Evaluations of LLMs consistently show that these models struggle with culturally grounded expressions such as idioms and idiomatic language, whose meanings depend on shared cultural and historical background that is not explicitly encoded in the linguistic form\cite{de-luca-fornaciari-etal-2024-hard, fu-etal-2025-chengyu}. These limitations directly affect translation quality: studies on interpretative slang and culture-specific items demonstrate that accurate translation often requires reconstructing intended meaning under context, rather than relying on direct lexical mapping\cite{liang2025slangditbenchmarkingllmsinterpretative, yao-etal-2024-benchmarking}.

Despite the growing awareness of these model limitations, current evaluation frameworks remain poorly equipped to detect or characterize culturally grounded translation errors. Standard automatic metrics such as BLEU and ChrF mainly reward surface lexical similarity, while learned metrics such as COMET often favor conventional phrasing and penalize stylistic variation, which makes them unreliable for evaluating culturally nuanced translations\cite{freitag-etal-2022-results, mukherjee-etal-2025-high}. As a result, figurative, cultural, and context dependent errors frequently go unnoticed. Translating idioms, resolving cultural ambiguity, or conveying implicit references across languages requires more than lexical fidelity, but these aspects are only weakly reflected in standard evaluation signals\cite{shafayat20252stepframeworkautomatedliterary, wicks2023identifyingcontextdependenttranslationsevaluation}. For instance, literal translations of idioms that preserve surface form but lose figurative meaning may still score highly under BLEU\cite{baziotis-etal-2023-automatic}. In response, recent work has proposed more context sensitive and culturally informed evaluation methods, including challenge sets that isolate cases requiring pragmatic reasoning or shared cultural knowledge, as well as metrics tailored to specific phenomena such as idiomaticity, metaphor, or cultural references\cite{SeaEval2023, yao-etal-2024-benchmarking}. Still, these advances remain limited in scope, since most evaluation datasets focus on isolated phenomena and lack a unified structure for taxonomy driven analysis of culturally induced meaning deviations in translation outputs.

To address these gaps, we introduce CulT-Eval, a benchmark designed to systematically evaluate machine translation performance on culture-loaded expressions. CulT-Eval provides curated source sentences and human reference translations, together with structured annotations that enable diagnostic analysis beyond sentence-level accuracy. Crucially, CulT-Eval is coupled with a unified error taxonomy that makes culturally induced meaning deviations explicit and measurable. Building on this taxonomy, we further propose a complementary evaluation metric that operationalizes these error categories and quantifies cultural meaning preservation beyond what standard automatic metrics can capture. Together, the benchmark and the metric form a coherent evaluation framework for analyzing how translation systems handle culturally grounded meaning. Using CulT-Eval, we conduct extensive evaluations of machine translation specialist systems and large language models, revealing systematic failure patterns that are not exposed by existing benchmarks or metrics.

\begin{itemize}
    
    \item We present CulT-Eval, a benchmark for evaluating machine translation of culture-loaded expressions, comprising over 7,900 carefully curated instances with structured coverage across diverse culturally grounded phenomena and diagnostic annotations.
    \item Through extensive evaluations of neural machine translation systems and large language models on CulT-Eval, we provide insights into recurring and systematic failure patterns in translating culturally grounded meaning, revealing challenges that are not exposed by existing benchmarks.
    \item We further find that widely used automatic evaluation metrics, such as BLEU and COMET, are insufficient for assessing culturally induced meaning deviations, motivating the use of a complementary, taxonomy-aware evaluation metric for more reliable analysis.
\end{itemize}

\section{Related Work}
\textbf{Culture-loaded Expressions in Machine Translation.}  Recent research has increasingly recognized the challenge of translating culture-loaded expressions, leading to the creation of numerous specialized evaluation benchmarks. Existing resources cover a wide range of culture-loaded content, spanning idioms\cite{li2023translatemeaningsjustwords} and proverbs\cite{wang2025proverbsrunpairsevaluating}, slang\cite{liang2025slangditbenchmarkingllmsinterpretative}and social-media expressions\cite{guo2025redefiningmachinetranslationsocial}, classical poetry\cite{chen-etal-2025-benchmarking-llms} and culture-specific items\cite{yao-etal-2024-benchmarking}, as well as domain-anchored terminology including recipes\cite{cao-etal-2024-cultural}. Despite this progress, existing resources remain fragmented along multiple axes, they are often bounded to a single cultural domain or register, focus on isolated linguistic phenomena, and employ incompatible annotation schemes and evaluation criteria, which limits cross-benchmark comparability and cumulative analysis of culture-related meaning shifts in MT. These limitations motivate the need for a unified taxonomy that can serve as a shared interface to characterize culture-loaded expressions and align evaluation across domains and phenomena.
\begin{table}
\centering
\small
\resizebox{\columnwidth}{!}{
\begin{tabular}{lcccccc}
\toprule
\multirow{2}{*}{\textbf{Category}} & \multirow{2}{*}{\textbf{Count}} & \multirow{2}{*}{\textbf{Ratio}} & \multicolumn{2}{c}{\textbf{Avg. Term Len. (Char)}} & \textbf{Avg. Context} \\
\cmidrule(lr){4-5}
 &  & (\%) & Source (Zh) & Target (En) & Source (Zh) \\
\midrule
Linguistic & 2,512 & 31.6 & 5.13 & 29.43 & 22.29 \\
Social & 2,399 & 30.1 & 3.21 & 24.22 & 23.72 \\
Material & 1,594 & 20.0 & 3.06 & 22.97 & 24.39 \\
Ecological & 833 & 10.5 & 2.82 & 20.13 & 22.66 \\
Religious & 621 & 7.8 & 3.13 & 23.01 & 24.22 \\
\midrule
\textbf{Total / Avg.} & \textbf{7,959} & \textbf{100.0} & \textbf{3.74} & \textbf{25.09} & \textbf{23.33} \\
\bottomrule
\end{tabular}
}
\caption{Statistics of the \textsc{CulT-Eval} benchmark.}
\label{tab:data_stats}
 \vspace*{-1em}
\end{table}
\noindent

\textbf{Evaluation Metrics in Machine Translation.}  Alongside fragmented resources, cultural evaluation is frequently conducted with general-purpose MT metrics that primarily capture surface overlap or holistic semantic similarity, including BLEU\cite{Papineni2002BleuAM}, ChrF++\cite{popovic-2017-chrf}, BERTScore\cite{Zhang2019BERTScoreET}, COMET\cite{Rei2020UnbabelsPI}, and QE-based variants\cite{Rei2022COMET22U2, Juraska2023MetricX23TG, Guerreiro2023xcometTM}. However, cultural meaning shifts often appear as dimension-specific errors, including culture-specific referent mismatch, loss of allusive meaning, and register or socio-pragmatic mismatch, which may not be reliably reflected by aggregate scores\cite{yang2025evaluatingllmschineseidiom, cheng2025seedxbuildingstrongmultilingual}. Overlap-based metrics can under-reward legitimate paraphrases or localization choices, while semantic-similarity and learned metrics may still fail to identify which cultural dimension is violated, limiting diagnostic value and impeding cross-benchmark interpretability\cite{proietti2025machinetranslationevaluationachieved, Tian2026BeyondLM}. These limitations highlight the need for evaluation that is taxonomy-grounded and dimension-aware, enabling consistent and fine-grained assessment of cultural meaning preservation across diverse benchmarks.

\section{CulT-Eval Benchmark}
\begin{figure*}[t]
    \centering
    \includegraphics[width=1\linewidth]{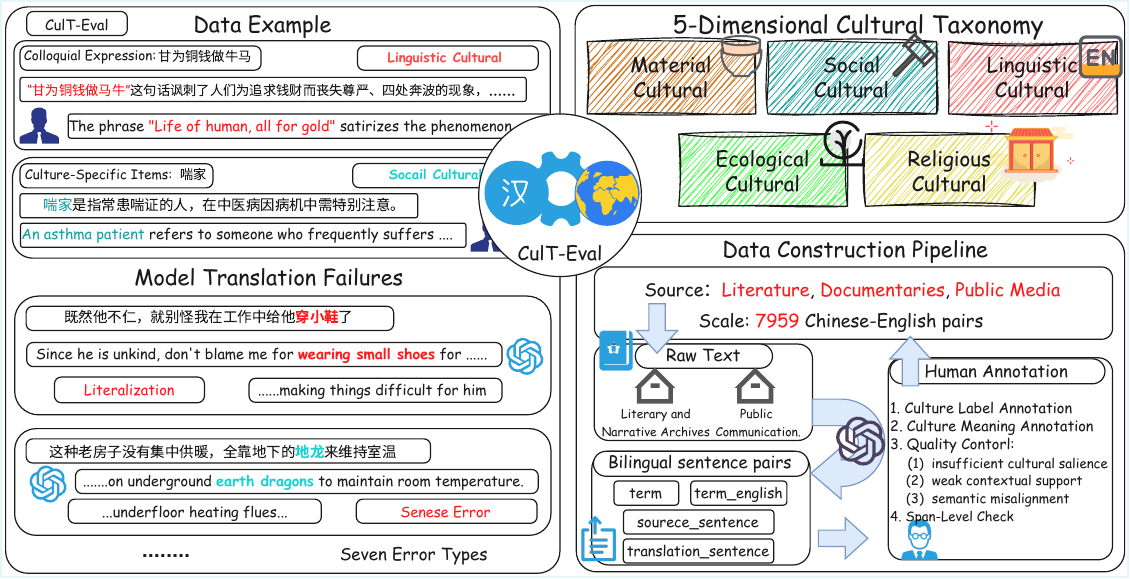}
    \caption{Overview of the CulT-Eval benchmark.}
    \label{fig:overview}
\end{figure*}

This section presents \textsc{CulT-Eval}, a benchmark designed to evaluate machine translation of culture-loaded expressions. We first delineate the data sources (§\ref{sec:data-acquisition}), followed by the process of expression identification and taxonomy-based classification (§\ref{sec:taxonomy}). We then detail the construction of human-verified references and the associated quality control procedures (§\ref{sec:construction}). An overview of the CulT-Eval is illustrated in Figure~\ref{fig:overview}.
\subsection{Data Source}
\label{sec:data-acquisition}
CulT-Eval is constructed from two major domains of culturally rich Chinese-English parallel data, selected to ensure diverse registers and translation ambiguity.
\paragraph{Literary and Narrative Archives.} We selected bilingual excerpts from regional literature, folklore chronicles, and movie subtitles. This domain captures the expressive richness of the language, serving as the primary source for idioms, slang, and Ecological/Material CSIs. By including both standardized idiomatic expressions and colloquial usage, this subset presents culturally nuanced phrasing that often requires interpretation beyond literal translation, and may or may not have established equivalents in the target language.
\paragraph{Public and Institutional Communication.} We aggregated data from official publicity materials, news reports, and documentaries. This category focuses on formal and communicative precision, providing a rich source for Social/Political CSIs and statutory terms. In contrast, the public/institutional subset focuses on consistent mappings of formalized terminology and policy-oriented expressions, which are typically expected to conform to existing bilingual conventions in cross-cultural communication.

\begin{table}[t]
    \centering
    \small
    \renewcommand{\arraystretch}{1.2}
    \setlength{\tabcolsep}{12pt} 
    \begin{tabular}{lcc}
        \toprule
        \textbf{Metric} & \textbf{Pearson $r$} & \textbf{Spearman $\rho$} \\
        \midrule
        BLEU         & 30.2 & 28.4 \\
        ChrF++       & 22.4 & 20.1 \\
        BERTScore    & 27.5 & 25.3 \\
        COMET        & 44.5 & 39.0 \\
        MetricX-QE   & 24.6 & 22.8 \\
        \midrule
        \textbf{ACRE (Ours)} & \textbf{68.4} & \textbf{65.1} \\
        \bottomrule
    \end{tabular}
    \caption{Pearson ($r$) and Spearman ($\rho$) correlation coefficients between metrics and human annotations on CulT-Eval.}
    \label{tab:metrics_corr}
    \vspace{-2em} 
\end{table}
\subsection{Cultural Taxonomy}
\label{sec:taxonomy}

\begin{CJK}{UTF8}{gbsn}
To facilitate fine-grained diagnostic evaluation, each instance in \textsc{CulT-Eval} is annotated with a cultural category. We adopt a five-way taxonomy adapted from established frameworks in translation studies and intercultural communication \cite{newmark1988approaches, Aixela+1999+52+78}. This taxonomy systematizes the underlying cultural grounding into the following five dimensions:

(1) \textbf{Material Culture}: Encompasses tangible artifacts, traditional attire, and architectural styles (e.g., \textit{Majiazi} [马架子], a log shelter). (2) \textbf{Social Culture}: Pertains to sociopolitical systems, historical movements, and institutional roles (e.g., \textit{Red Tourism} [红色旅游]). (3) \textbf{Linguistic Culture}: Covers idiomatic expressions, proverbs, and metaphors with non-compositional meanings (e.g., \textit{Three cobblers} [三个臭皮匠]). (4) \textbf{Religious Culture}: Refers to belief systems, ritualistic practices, and philosophical frameworks (e.g., Confucianism). (5) \textbf{Ecological Culture}: Includes terms rooted in seasonal cycles, calendrical systems, and geography-based cosmological concepts (e.g., \textit{Grain Rain} [谷雨]). Each instance is assigned a single primary label based on its predominant contextual function. To ensure diagnostic clarity, we prioritize mutually exclusive assignments even for expressions that exhibit categorical overlap.
\end{CJK}

\subsection{Benchmark Construction Pipeline}
\label{sec:construction}

The construction of CulT-Eval followed a semi-automated pipeline combining LLM assistance with human annotation.
\paragraph{LLM-Assisted Candidate Extraction.} We employed GPT-5 to identify candidate sentences from raw Chinese corpora likely to contain culture-loaded expressions. Source texts spanned the domains, including literary works, documentaries, movie subtitles. The model was prompted with domain-adapted instructions (see Table~\ref{fig:prompt_mining}) to overgenerate potentially culture-specific content, prioritizing recall over precision. 
\paragraph{Human Annotation and Cultural Term Labeling.}
All GPT-extracted candidates were reviewed by trained annotators to verify the presence of at least one genuine culture-loaded expression. Sentences deemed culturally irrelevant or ambiguous were excluded. For each retained sentence, annotators identified the minimal span of the culture-specific term in both the Chinese source and the English target, and assigned a category label from the taxonomy defined in Section~\ref{sec:taxonomy}.

In addition to span identification and classification, annotators enriched each instance with: (1) a verified English reference translation drawn from the original bilingual source; (2) a cultural explication, a one-sentence contextual definition that explains the term’s cultural or historical significance; and (3) a standardized English equivalent when applicable.
\begin{table*}[!t]
\centering
\small
\setlength{\tabcolsep}{5.5pt} 
\renewcommand{\arraystretch}{1.2} 
\emergencystretch=1em

\begin{tabular}{
l | cc || cc || cc || cc || cc
}
\toprule
\multirow{2}{*}{\textbf{Model}} & 
\multicolumn{2}{c||}{\textbf{BLEU}} & 
\multicolumn{2}{c||}{\textbf{CHRF++}} & 
\multicolumn{2}{c||}{\textbf{BERTScore}} & 
\multicolumn{2}{c||}{\textbf{COMET}} &
\multicolumn{2}{c}{\textbf{MetricX-QE}} \\ 

\cmidrule(lr){2-3} \cmidrule(lr){4-5} \cmidrule(lr){6-7} \cmidrule(lr){8-9} \cmidrule(lr){10-11}

& 0-shot & 1-shot 
& 0-shot & 1-shot 
& 0-shot & 1-shot 
& 0-shot & 1-shot 
& 0-shot & 1-shot \\ 
\midrule

\multicolumn{11}{c}{\textit{\textbf{Machine Translation Models}}}\\
\midrule
Hunyuan-MT-7B       & -- & -- & -- & -- & -- & -- & -- & -- & -- & -- \\
Madlad400-10B-MT    & -- & -- & -- & -- & -- & -- & -- & -- & -- & -- \\
NLLB200-3-3B        & -- & -- & -- & -- & -- & -- & -- & -- & -- & -- \\

\midrule
\multicolumn{11}{c}{\textit{\textbf{Open-Sourced and Proprietary LLMs}}}\\
\midrule
Llama-3.1-8B-Instruct & -- & -- & -- & -- & -- & -- & -- & -- & -- & -- \\
DS-R1-D-Qwen-7B & -- & -- & -- & -- & -- & -- & -- & -- & -- & -- \\
Qwen3-8B-Instruct     & -- & -- & -- & -- & -- & -- & -- & -- & -- & -- \\
Qwen3-32B-Instruct    & -- & -- & -- & -- & -- & -- & -- & -- & -- & -- \\
DeepSeek-v3         & -- & -- & -- & -- & -- & -- & -- & -- & -- & -- \\
GPT-5.1             & -- & -- & -- & -- & -- & -- & -- & -- & -- & -- \\

\bottomrule
\end{tabular}

\caption{Comprehensive evaluation results across five metrics.}
\label{tab:translation_results}
\end{table*}
\subsection{Dataset Statistics and Quality Control}
\label{sec:stats}

After annotation, we applied post hoc filtering to ensure consistency and interpretability of the benchmark. Annotators refined span boundaries, normalized terminology, and excluded instances that did not meet the dataset criteria. Specifically, instances were removed if they exhibited: (1) insufficient cultural salience, where the expression did not encode a clear culture-dependent concept; (2) weak contextual support, where the surrounding sentence was insufficient to disambiguate meaning; or (3) semantic misalignment, where the source and target sentences showed low correspondence due to overly literal translation or structural mismatch.

English reference translations were drawn from the original bilingual sources, including subtitles, literary translations, and official publications. These translations were manually inspected to verify basic pragmatic adequacy. Instances were excluded if the English side consisted primarily of unglossed transliterations or if the translation failed to reflect the intended meaning of the culture-loaded expression.

In addition to sentence-level alignment, we annotated the corresponding spans of culture-loaded expressions on the English reference side. Each instance therefore contains an explicit mapping between the source-language cultural span and its realization in the target language, enabling span-level inspection during evaluation.

From a linguistic perspective, the dataset encompasses culture-loaded expressions realized as idioms, slang, and colloquialisms, alongside literary and poetic forms and entity-based culture-specific items. After rigorous filtering, the dataset was refined from an initial pool of approximately 12,000 candidates to (see Table~\ref{tab:data_stats} for detailed statistics).

\section{Evaluation and Metric Analysis}
\label{4.1}

This section evaluates translation models on CulT-Eval and analyzes the adequacy of commonly used evaluation metrics for culture-loaded translation. We begin with standard sentence-level metrics, and then examine their behavior on culturally salient spans. 
\subsection{Sentence-Level Evaluation with Standard Metrics}
\label{sec:standard-metrics}
We evaluate a set of representative translation systems on CulT-Eval, including both multilingual NMT models and instruction-tuned large language models (LLMs) used in a zero-shot translation setting. The MT baselines include publicly released models such as NLLB-200-3.3B\cite{nllbteam2022languageleftbehindscaling}, hunyuan-7B\cite{zheng2025hunyuanmttechnicalreport} and MADLAD-400-10B\cite{kudugunta2023madlad400multilingualdocumentlevellarge}. In addition, we evaluate several LLM-based systems, including GPT-5.1 \cite{singh2025openaigpt5card}, Llama-3.1-Instruct \cite{grattafiori2024llama3herdmodels}, and Qwen3-Instruct series \cite{yang2025qwen3technicalreport} and DeepSeek-V3 \cite{deepseekai2025deepseekv3technicalreport}. All systems are evaluated in a source-only setting, where models generate English translations directly from Chinese source sentences. For LLMs, we employ two prompting paradigms: (i) a vanilla zero-shot translation prompt, and (ii) a one-shot prompt containing a single illustrative example. To ensure a rigorous evaluation, the one-shot example is held-out from the test set, and no supplementary cultural explications or reference translations are provided at inference time. NMT systems are evaluated under their standard inference settings without prompt variations.

    

\subsection{Sentence-level Metrics under Cultural Evaluation}
\label{4.2}
In Table~\ref{tab:translation_results},sentence-level metrics distinguish translation systems on CulT-Eval, with both NMT and LLM-based models achieving competitive scores and clear performance differences. However, these metrics do not explicitly evaluate whether culturally salient content is correctly preserved, which is central to the CulT-Eval task.

To assess whether sentence-level metrics reflect cultural correctness, we analyze their alignment with human judgments, defined as whether the annotated culture-loaded span is correctly expressed in the translation. Table~\ref{tab:metrics_corr} reports segment-level correlations between automatic metrics and human judgments of cultural correctness. Across metrics, BLEU, ChrF, BLEURT, and COMET exhibit weak and unstable correlations, indicating that higher sentence-level scores do not reliably correspond to correct translation of culture-loaded spans.

\begin{CJK*}{UTF8}{gbsn}

\begin{table}[h] 
    \centering

    \small 
    
    \renewcommand{\arraystretch}{1.4} 
    
    \begin{tabularx}{\linewidth}{@{}l X c c@{}}
        \toprule
        \textbf{Type} & \textbf{Content Analysis} & \textbf{Score} & \textbf{Valid} \\
        \midrule
        
        \makecell[l]{High \\ \textit{Error}} & 
        \lbl{SRC} \dots 强调自己\textbf{吃软不吃硬}，如果\dots \newline
        \lbl{REF} \textit{\dots amenable to coaxing but not coercion\dots} \newline
        \lbl{SYS} \dots better to persuasion than force\dots 
        & \makecell{32.1 \\ 0.83} & \xmark \\
        
        \cmidrule(lr){1-4} 

        \makecell[l]{Low \\ \textit{Good}} & 
        \lbl{SRC} \dots 典型的\textbf{关系户}问题，在一些\dots \newline
        \lbl{REF} \textit{\dots typical issue of nepotism, which\dots} \newline
        \lbl{SYS} \dots typical case of nepotism driven by\dots
        & \makecell{20.3 \\ 0.64} & \cmark \\
        
        \bottomrule
    \end{tabularx}
    

    \raggedright 
    \scriptsize \textit{* Score: BLEU (top) / COMET (bottom).}
        \caption{Performance Analysis.}
    \label{tab:case_study_adaptive}
    \vspace*{-2.5em}
\end{table}

\end{CJK*}
Representative examples in Table~\ref{tab:case_study_adaptive} further illustrate this misalignment. We observe translations that receive high sentence-level scores despite literalization, over-generalization, or omission of cultural meaning, as well as translations that accurately convey cultural meaning through paraphrasing or explicitation but receive low scores due to surface-level divergence from the reference.

Together, these results demonstrate that sentence-level similarity is an unreliable proxy for cultural correctness. While standard metrics capture overall translation quality, they fail to reflect whether culturally salient meaning is preserved, motivating a more fine-grained analysis of cultural translation errors.

\subsection{A Taxonomy of Culture-related Translation Errors}
Beyond the empirical observations in Section~\ref{4.2}, which suggest a systematic misalignment between metrics and cultural correctness, we conduct a structured error analysis to characterize the distortion of cultural meaning. We introduce a taxonomy of culture-related translation errors to formalize these failure modes. By distinguishing specific types of cultural attrition, our taxonomy provides a rigorous framework for evaluating translation quality where conventional metrics fail.
\begin{figure}
    \centering
    \includegraphics[width=1\linewidth]{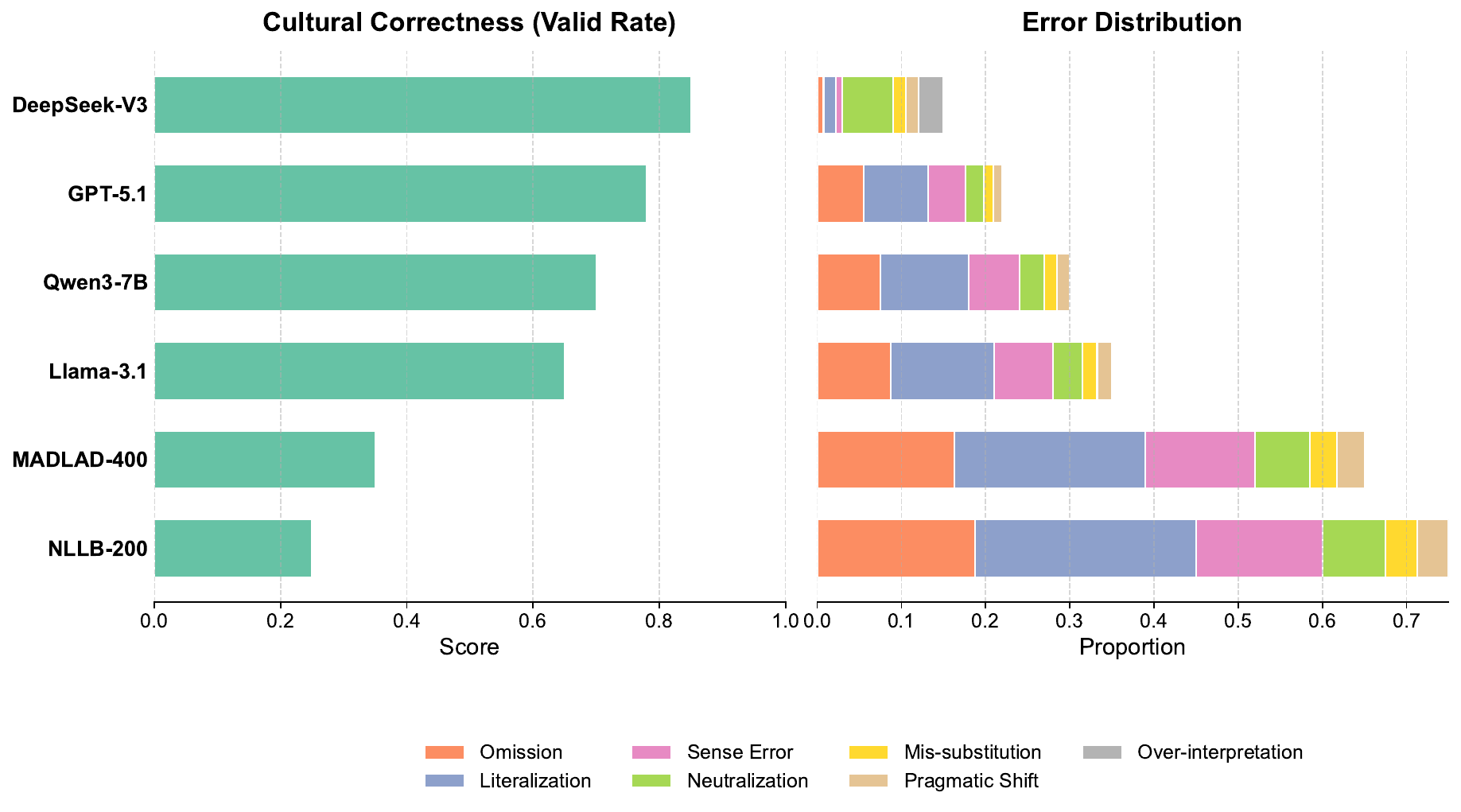}
    \caption{Performance analysis of six selected models. The left chart displays the overall Cultural Correctness score. The right chart visualizes the distribution of seven specific error types within the incorrect samples.}
    \label{error categories count}
    \vspace*{-1em}
\end{figure}
\subsubsection{Annotation Principle}
A key challenge in analyzing cultural translation errors is that multiple error phenomena may co-occur within a single instance. To ensure consistency, we assign a primary error label according to a fixed priority order, ranging from omission to over-interpretation. This ordering reflects a progression from the complete absence of cultural realization to distorted or excessive realization, and ensures that each instance is associated with the most fundamental source of cultural failure. Detailed definitions and examples are provided in the Appendix~\ref{}.

\subsubsection{Error Categories.}
We identify seven recurrent error categories that capture distinct ways in which cultural meaning may fail to be correctly realized in translation. All categories refer specifically to errors in the translation of culture-loaded spans, rather than to general translation errors.(1) \textbf{Omission:} the culture-loaded span is not realized in the translation, either through deletion or replacement with an empty or vacuous expression.  (2)\textbf{Literalization: } the translation preserves surface meaning through word-by-word rendering but fails to activate the conventional or idiomatic cultural sense. (3)
\textbf{Sense Error:} an incorrect sense or referent is selected, resulting from misinterpretation rather than deliberate cultural substitution. (4) \textbf{Neutralization:} the functional meaning is broadly conveyed, but culturally specific features are flattened into generic expressions, weakening cultural salience.
(5) \textbf{Mis-substitution:} a target-culture analogue is used as a replacement, but the analogy is misleading or non-equivalent. (6) \textbf{Pragmatic Shift: } social or interactional meaning is altered, including changes in politeness, honorifics, or perceived social relations. (7) \textbf{Over-interpretation:} additional cultural explanations, background information, or value judgments are introduced beyond what is explicit in the source.

\subsubsection{Error Distributions across Systems}
To illustrate how the proposed taxonomy manifests in practice, we examine the distribution of cultural correctness and error types across a set of representative translation systems, all evaluated under the same source-only setting described in Section~\ref{4.1}.

Figure~\ref{error categories count} presents the cultural correctness rate and the composition of primary error categories for each system. The results show that cultural failures are systematic and vary in composition across systems, even when overall translation quality at the sentence level appears competitive. Importantly, these differences arise from how culturally loaded spans are realized, such as through omission, literalization, neutralization, pragmatic distortion, or overinterpretation, rather than from fluency or grammaticality on the surface.

\begin{table}[h]
    \centering

    \small 
    
    \setlength{\tabcolsep}{2pt} 
    
    \renewcommand{\arraystretch}{1.3} 
    
    \begin{tabularx}{\linewidth}{@{} X c c @{\hskip 4pt} c c @{}}
        \toprule
        \multirow{2}{*}{\textbf{Error Type}} & \multicolumn{2}{c}{\textbf{Lexical}} & \multicolumn{2}{c}{\textbf{Semantic}} \\
        \cmidrule(r){2-3} \cmidrule(l){4-5} 
        
         & BLEU & ChrF & BERT & COMET \\
        \midrule
        
         Omission            & \yes & \yes & \yes & \yes \\
         Literalization      & \no  & \no  & \no  & \no  \\
         Neutralization      & \no  & \no  & \partialyes & \partialyes \\
         Over-interpretation & \no  & \no  & \no  & \no  \\
        
        \bottomrule
    \end{tabularx}
    
    \caption{Metric sensitivity analysis. (\yes: Sensitive; \no: Insensitive; \partialyes: Partial)}
    \label{tab:error_sensitivity_large}
    \vspace*{-2em}
\end{table}

\subsection{Structural Limitations of Sentence-level Evaluation for Cultural Correctness}

Taken together, the analyses reveal a structural mismatch between sentence-level evaluation and cultural correctness. Sentence-level metrics assume that overall similarity to a reference reflects meaning preservation, an assumption that breaks down when culturally salient meaning is realized through specific spans that may not substantially affect surface form.

The error taxonomy makes this mismatch explicit by distinguishing failure modes with different interactions with surface similarity. While omission errors remove content and are therefore often penalized, errors such as literalization, neutralization, and over-interpretation frequently preserve lexical or semantic overlap while distorting culturally salient meaning, allowing affected translations to receive high sentence-level scores.

Table~\ref{tab:error_sensitivity_large} shows that this asymmetry is systematic. N-gram–based metrics consistently penalize omission but are largely insensitive to error types that preserve surface overlap, while embedding-based metrics exhibit only limited improvements and remain unreliable for detecting several prevalent cultural error categories.

These findings indicate that the limitation of sentence-level metrics stems from the evaluation objective itself rather than from metric design. By aggregating similarity at the sentence level, different cultural failure mechanisms are conflated into a single score, motivating the need for an evaluation approach that explicitly targets the realization of culture-loaded spans. In the next section, we introduce ACRE to operationalize this perspective.

\section{ACRE}
We propose ACRE (\textbf{A}nchored \textbf{C}ultural \textbf{R}ealization \textbf{E}valuation), an automatic evaluation metric for assessing whether culturally salient meaning is correctly realized in translation. Unlike sentence-level similarity metrics that rely on surface overlap, ACRE is explicitly anchored by \textbf{Cultural Explication} annotations, which provide ground-truth definitions of culture-loaded spans and frame cultural evaluation as a structured verification problem rather than open-ended judgment.

ACRE models cultural realization through two complementary components: \textbf{Validity} and \textbf{Quality}. Validity determines whether the intended cultural referent is correctly instantiated, while Quality assesses how appropriately that meaning is expressed. Formally, let \(S\) denote the source sentence, \(H\) the hypothesis translation, \(E\) the associated Cultural Explication, and \(C\) the category of the culture-loaded span. ACRE is defined as:
\begin{equation}
\label{eq:acre}
\mathrm{ACRE}(S, H, E, C)
=
\mathbb{I}_{\mathrm{valid}}(H, E)
\cdot
\Phi_{\mathrm{quality}}(H, S, C)
\end{equation}

Here, \(\mathbb{I}_{\mathrm{valid}}(H, E) \in \{0, 1\}\) is the \textbf{Semantic Validity Indicator}, computed by the \textbf{Semantic Validator}, which verifies whether the hypothesis instantiates the cultural referent defined in \(E\). Translations that fail this check are assigned a score of zero, preventing fluent but semantically incorrect realizations from being rewarded.

The Quality component \(\Phi_{\mathrm{quality}}\) is computed only for valid instances:
\begin{equation}
\label{eq:quality}
\Phi_{\mathrm{quality}}(H, S, C)
=
\alpha_{C} \cdot \mathcal{S}_{\mathrm{fidelity}}(H, S)
+
\beta_{C} \cdot \mathcal{S}_{\mathrm{clarity}}(H)
\end{equation}
where \(\mathcal{S}_{\mathrm{fidelity}}\) is the \textbf{Fidelity Score}, assessing the preservation of intended meaning and pragmatic force, and \(\mathcal{S}_{\mathrm{clarity}}\) is the \textbf{Clarity Score}, assessing communicative intelligibility for target-language readers.

\subsection{Category-conditioned Evaluation and Reference Implementation}

Based on the taxonomy introduced in Section~\ref{4.1}, ACRE adopts category-conditioned protocols that determine how Quality is assessed, while leaving the metric definition unchanged. A Category Check first routes each instance to one of two evaluation protocols according to its category \(C\).

For Protocol A (Fact-centric), which applies to categories such as specific cultural concepts, material artifacts, and social institutions, evaluation emphasizes referential validity. Under this protocol, strict alignment with the Cultural Explication is required, and deviations in referential identity directly invalidate the translation via the Semantic Validator.

For Protocol B (Style-centric), which applies to idioms, literary expressions, and slang, evaluation emphasizes functional equivalence. Under this protocol, paraphrasing or re-expression is permitted as long as pragmatic force and register are preserved.

\begin{figure}
    \centering
    \includegraphics[width=1\linewidth]{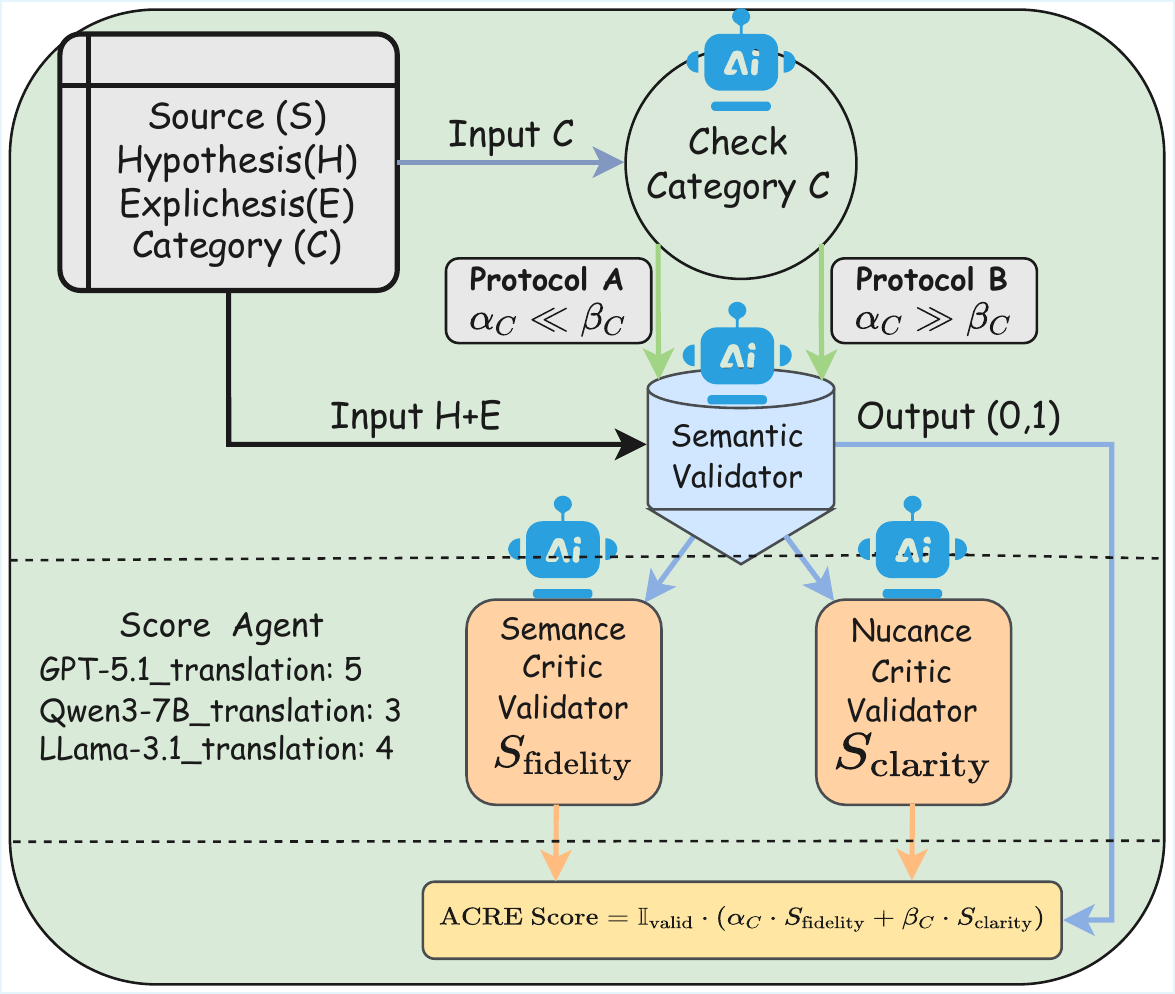}
    \caption{Evaluation pipeline}
    \label{fig:evaluation}
    \vspace*{-1em}
\end{figure}

These category-conditioned protocols are instantiated in a multi-agent evaluation framework, termed CulT-Agent, as illustrated in Figure~\ref{fig:evaluation}. The framework operationalizes the Validity and Quality components of ACRE through coordinated agents that correspond directly to the metric formulation.

Within CulT-Agent, Validity is assessed by the Semantic Validator. For instances that pass Validity, Quality is assessed by two complementary agents: the Fidelity Critic, which computes the Fidelity Score, and the Nuance Critic, which computes the Clarity Score. These agents implement the Quality function defined in Eq.~(\ref{eq:quality}).

Throughout evaluation, reference translations are used only as stylistic anchors rather than semantic ground truth, while Cultural Explications serve as the authoritative basis for determining cultural correctness. Detailed descriptions of individual agents and their coordination protocols are provided in Appendix~\ref{app:agents}.

\subsection{Implementation Details}

In our experiments, ACRE is instantiated through CulT-Agent, which realizes the Semantic Validator, Fidelity Critic, and Nuance Critic using a large language model as a constrained judge. Each agent executes a specific functional role defined by the category-conditioned protocols.

All evaluations are implemented with grok-4.1-fast, using deterministic decoding and fixed prompts. The prompts instantiate the roles of individual agents and their coordination protocol, and are held constant across all experiments to ensure reproducibility. For transparency, all prompt templates and agent specifications are provided in Appendix~\ref{sec:prompt}
\section{Experiments and Analyses}
\subsection{Cultural Correctness under ACRE}
\label{6.1}
We evaluate ACRE using the same set of~\ref{sec:standard-metrics} translation systems and experimental configuration described in Section. Table~\ref{tab:acre_full_results} reports ACRE scores and their components for representative machine translation systems and large language models on CulT-Eval.
\begin{table*}[!t]
\centering
\small
\setlength{\tabcolsep}{8.5pt} 
\renewcommand{\arraystretch}{1.2}
\emergencystretch=1em

\begin{tabular}{
l | cc || cc || cc
}
\toprule
\multirow{2}{*}{\textbf{Model}} & 
\multicolumn{2}{c||}{\textbf{Stage I: Validity}} & 
\multicolumn{2}{c||}{\textbf{Stage II: Quality}} & 
\multicolumn{2}{c}{\textbf{Final ACRE}} \\

\cmidrule(lr){2-3} \cmidrule(lr){4-5} \cmidrule(lr){6-7}

& 0-shot & 1-shot & Fidelity & Clarity & 0-shot & 1-shot \\ 
\midrule

\multicolumn{7}{c}{\textit{\textbf{Machine Translation Models}}}\\
\midrule
Hunyuan-MT-7B         & -- & -- & -- & -- & -- & -- \\
Madlad400-10B-MT      & -- & -- & -- & -- & -- & -- \\
NLLB200-3-3B          & -- & -- & -- & -- & -- & -- \\

\midrule
\multicolumn{7}{c}{\textit{\textbf{Open-Sourced and Proprietary LLMs}}}\\
\midrule
Llama-3.1-8B-Instruct & -- & -- & -- & -- & -- & -- \\
DS-R1-D-Qwen-7B       & -- & -- & -- & -- & -- & -- \\
Qwen3-8B-Instruct     & -- & -- & -- & -- & -- & -- \\
Qwen3-32B-Instruct    & -- & -- & -- & -- & -- & -- \\
DeepSeek-v3           & -- & -- & -- & -- & -- & -- \\
GPT-5.1               & -- & -- & -- & -- & -- & -- \\

\bottomrule
\end{tabular}

\caption{Comprehensive ACRE evaluation results across proprietary and open-source models.}
\label{tab:acre_full_results}
\end{table*}

The results indicate that the realization of culture-loaded expressions remains a challenging and unreliable aspect of current translation systems. Across models, validity aspect of current translation systems. Across models, validity failures are frequent, suggesting that culture-loaded expressions are often misinterpreted, generalized, or omitted, even when translation appear fluent at the sentence level. Such failures directly limit final ACRE scores and reveal deficiencies that are obscured by sentence-level similarity metrics.

While stronger language models achieve higher validity rates than machine translation systems, these improvements are still incremental rather than defintive. Correct handling of culture-loaded expressions is not consistently guaranteed, and substantial variation persists in how cultural meaning is preserved and expressed once referential correctness is satisfied. Importantly, these differences are not driven by surface-level similarity or grammaticality, but by how systems handle culture-loaded expressions at both the referential and pragmatic levels. By decomposing cultural correctness into semantic validity and realization quality, ACRE provides a more informative view of translation behavior than aggregate sentence-level metrics, which conflate distinct sources of cultural failure.
\begin{table}[t]
\centering
\small
\begin{tabular}{l|c|c}
\toprule
\textbf{Configuration} & \textbf{Correlation ($r$)} & \textbf{$\Delta r$} \\
\midrule
\textbf{Full ACRE} & \textbf{0.88} & -- \\
\midrule
\textit{w/o} Validity Gate ($\mathbb{I}_{\text{valid}}$) & 0.76 & -0.12 \\
\textit{w/o} Adaptive Routing (Protocols) & 0.81 & -0.07 \\
\textit{w/o} Explication Anchor ($E$) & 0.62 & \textbf{-0.26} \\
\textit{w/o} Reference Anchor ($R$) & 0.85 & -0.03 \\
\bottomrule
\end{tabular}
\caption{Ablation study of ACRE. We compare the semantic validity (Stage I) and quality profiling (Stage II) across zero-shot and one-shot settings.}
\label{tab:ablation}
\vspace*{-1em}
\end{table}

\subsection{Ablation Study}
To verify the contribution of each component in ACRE, we conduct an ablation study on the translations generated by Qwen3-8B. Table~\ref{tab:ablation} reports the correlation results. Removing the semantic validity gate leads to a drop in human alignment, confirming its role in filtering hallucinated but fluent translations. The most significant degradation occurs when excluding Cultural Explications, demonstrating that external semantic grounding is more critical for cultural evaluation than internal parametric knowledge alone. In contrast, the absence of reference translations causes only a marginal performance decline, indicating that ACRE relies more on semantic definitions than on surface-form similarity. These results underscore that ACRE’s components provide complementary diagnostic power, with explications serving as the primary anchor for cultural correctness.

\subsection{Diagnostic Sensitivity to Cultural Error Types}

\begin{figure}[t]
    \centering
    \includegraphics[width=1\linewidth]{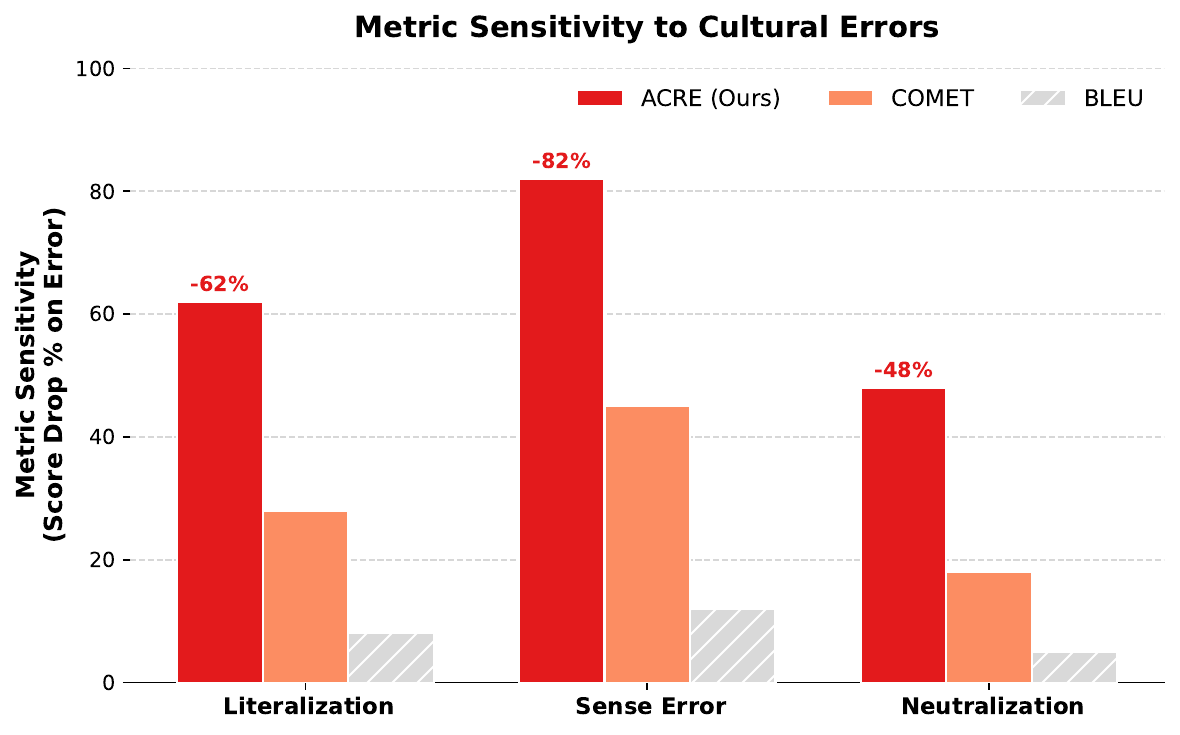}
    \caption{Sensitivity of evaluation metrics to cultural translation errors.}
    \label{comparison}
    \vspace*{-1em}
\end{figure}

Figure~\ref{comparison} compares the sensitivity of ACRE and sentence-level metrics to representative cultural error types in the translation of \textit{culture-loaded expressions}. ACRE exhibits substantial score drops for sense errors, literalization, and neutralization, indicating strong responsiveness to failures that distort cultural meaning while preserving surface overlap. In contrast, COMET shows only moderate sensitivity, and BLEU remains largely insensitive across all error types, with minimal score variation even when cultural meaning is severely distorted. These results demonstrate that ACRE captures diagnostic signals that are systematically missed by sentence-level similarity metrics, directly explaining the discrepancies observed in Section~\ref{6.1}.

\section{Conclusion}
In this paper, we introduced CulT-Eval, a large-scale benchmark for evaluating machine translation of culture-loaded expressions, together with a unified cultural taxonomy and fine-grained error annotations. Through extensive evaluation, we showed that widely used sentence-level metrics fail to reliably reflect cultural correctness, often overlooking systematic meaning distortions. To address this gap, we proposed ACRE, a taxonomy-aware evaluation metric anchored in cultural explications, which demonstrates substantially stronger alignment with human judgments and higher diagnostic sensitivity to culture-related errors. Our findings highlight the limitations of surface-level evaluation and underscore the need for culturally grounded assessment frameworks. We hope CulT-Eval and ACRE will facilitate more reliable evaluation and foster future research on culturally aware machine translation.

\section{Acknowledgments}

Identification of funding sources and other support, and thanks to
individuals and groups that assisted in the research and the
preparation of the work should be included in an acknowledgment
section, which is placed just before the reference section in your
document.

This section has a special environment:
\begin{verbatim}
  \begin{acks}
  ...
  \end{acks}
\end{verbatim}
so that the information contained therein can be more easily collected
during the article metadata extraction phase, and to ensure
consistency in the spelling of the section heading.

Authors should not prepare this section as a numbered or unnumbered {\verb|\section|}; please use the ``{\verb|acks|}'' environment.

\bibliographystyle{ACM-Reference-Format}
\bibliography{sample-base}

\appendix

\section{Prompt}
\label{sec:prompt}

\begin{figure*}[h]
\begin{tcolorbox}[
  title={Prompt Template: Culture-Mining Agent (Data Extraction)},
  colframe=black, colback=white, coltext=black,
  boxrule=0.4pt, arc=1.5pt,
  left=3pt,right=3pt,top=3pt,bottom=3pt,
  boxsep=2pt, toptitle=2pt,bottomtitle=2pt,
  fonttitle=\bfseries\small, fontupper=\small
]
\textbf{System Instruction:}
You are a bilingual data mining expert. Your task is to extract "Culture-Loaded Sentence Pairs" from the provided raw bilingual text.

IMPORTANT CONSTRAINTS:
\begin{itemize}\setlength\itemsep{1pt}
  \item You must identify sentences containing Idioms, Slang, or Culture-Specific Items (e.g., history, food, traditions).
  \item You must extract both the Chinese source and the English translation.
  \item Ignore common sentences that lack specific cultural depth.
  \item Output must be a strictly valid JSON list.
\end{itemize}

\textbf{User Input:}\par
{\ttfamily\small
[User]
Extract cultural pairs from the following text chunk:

Raw Text:
\{RAW\_TEXT\_CHUNK\}

Return format:
[
  \{
    "src": "Chinese sentence...",
    "tgt": "English sentence...",
    "focus\_term": "The specific cultural word"
  \}
]
}
\end{tcolorbox}
\caption{The prompt template for the Culture-Mining.}
\label{fig:prompt_mining}
\end{figure*}

\begin{CJK}{UTF8}{gbsn}
\begin{figure*}[h]
\begin{tcolorbox}[
  title={Prompt Template: Fine-grained Taxonomy Classifier},
  colframe=black, colback=white, coltext=black,
  boxrule=0.4pt, arc=1.5pt,
  left=3pt,right=3pt,top=3pt,bottom=3pt,
  boxsep=2pt, toptitle=2pt,bottomtitle=2pt,
  fonttitle=\bfseries\small, fontupper=\small
]
\textbf{System Instruction:}
You are a Cultural Linguist. Your task is to classify a specific Chinese term into one of the \textbf{Five Cultural Categories}.

\textbf{Taxonomy Definitions:}
\begin{enumerate}\setlength\itemsep{0pt}

  \item \textbf{Ecological Culture:} Terms related to specific animals, plants, geography, climate, or natural phenomena unique to the region (e.g., 梅雨, 熊猫, 黄河).
  \item \textbf{Material Culture:} Terms related to food, clothing, architecture, artifacts, or daily necessities (e.g., 旗袍, 饺子, 炕, 四合院).
  \item \textbf{Social Culture:} Terms related to institutions, titles, festivals, customs, history, or social hierarchy (e.g., 高考, 春节, 尚书, 关系户).
  \item \textbf{Religious Culture:} Terms related to beliefs, mythology, philosophy (Confucianism, Taoism, Buddhism), or taboos (e.g., 阴阳, 菩萨, 玉帝).
  \item \textbf{Linguistic Culture:} Terms involving idioms (Chengyu), metaphors, slang, proverbs, or witticisms (e.g., 吃软不吃硬, 摸鱼, 破防).
\end{enumerate}
\textbf{User Input:}\par
{\ttfamily\small
[User]
Context Sentence: \{SOURCE\_SENTENCE\}
Target Term: \{FOCUS\_TERM\}

Task:
1. Analyze the meaning of the term in context.
2. Assign strictly ONE category from the list above.
3. Provide a short explanation.

Return Format JSON:
\{
  "term": "\{FOCUS\_TERM\}",
  "category": "Social Culture",
  "reason": "It refers to a specific historical government position."
\}
}
\end{tcolorbox}
\caption{Prompt template for fine-grained cultural classification.}
\label{fig:prompt_taxonomy}
\end{figure*}

\end{CJK}

\begin{figure*}[t]
\begin{tcolorbox}[
  title={Prompt Template: Core Dispatcher (Protocol Routing)},
  colframe=black, colback=white, coltext=black,
  boxrule=0.4pt, arc=1.5pt,
  left=3pt,right=3pt,top=3pt,bottom=3pt,
  boxsep=2pt, toptitle=2pt,bottomtitle=2pt,
  fonttitle=\bfseries\small, fontupper=\small
]
\textbf{System Instruction:}
You are a taxonomy expert in cultural linguistics. Your task is to classify the "Cultural Category" of a specific term within a source sentence.

IMPORTANT DEFINITIONS:
\begin{itemize}\setlength\itemsep{0pt}
  \item \textbf{Protocol A (Fact-Centric):}
    \begin{itemize}
      \item Specific Concepts: Concrete institutions, historical artifacts, unique objects, technical items.
      \item Proper Nouns: Names of unique places, people, organizations, festivals.
      \item \textbf{Core Logic:} Referential Precision. The translation must point to the EXACT same entity.
      \item \textbf{Tie-Breaker:} If a term has a specific history/material existence, choose A (e.g., "Forbidden City").
    \end{itemize}
  \item \textbf{Protocol B (Style-Centric):}
    \begin{itemize}
      \item Figurative Language: Idioms (Chengyu), metaphors, allegories.
      \item Slang/Pop Culture: Buzzwords, internet memes, dialect words used for effect.
      \item \textbf{Core Logic:} Pragmatic Equivalence. Imagery and tone are more important than literal words.
      \item \textbf{Tie-Breaker:} If the term describes a situation/feeling rather than an object, choose B (e.g., "eating vinegar").
    \end{itemize}
\end{itemize}

\textbf{User Input:}\par
{\ttfamily\small
[User]
Analyze the cultural term inside the brackets "[]" in the source sentence.

[Input]
Source: \{SOURCE\_WITH\_BRACKETS\}

[Task]
Classify the term based on its primary function in this specific context.
Return ONLY one label: "Protocol A" or "Protocol B".
}
\end{tcolorbox}
\caption{The Core Dispatcher prompt. It routes instances to the correct evaluation protocol (Fact-Centric or Style-Centric) based on linguistic features and functional context.}
\label{prompt:dispatcher}
\end{figure*}

\begin{figure*}[t]
\begin{tcolorbox}[
  title={Prompt Template: Stage I - Semantic Validator (The Gate)},
  colframe=black, colback=white, coltext=black,
  boxrule=0.4pt, arc=1.5pt,
  left=3pt,right=3pt,top=3pt,bottom=3pt,
  boxsep=2pt, toptitle=2pt,bottomtitle=2pt,
  fonttitle=\bfseries\small, fontupper=\small
]
\textbf{System Instruction:}
You are a strict Semantic Validator. Your ONLY goal is to detect "Hallucinations" (Error Type B3) or "Severe Mis-substitutions" (Error Type B5).

CRITICAL GROUND TRUTH RULE:
\begin{itemize}\setlength\itemsep{0pt}
  \item You must rely EXCLUSIVELY on the provided [Cultural Explication].
  \item If the Explication says X, and the Model translates it as Y (where Y!=X), it is INVALID.
  \item Ignore your own internal knowledge if it conflicts with the Explication.
\end{itemize}

NEGATIVE CONSTRAINTS (What NOT to check):
\begin{itemize}\setlength\itemsep{0pt}
  \item DO NOT check for fluency, grammar, or style.
  \item DO NOT check for "neutralization" (loss of flavor). A boring but factually correct translation is VALID.
  \item DO NOT penalize literal translations here, provided they refer to the correct concepts.
\end{itemize}

\textbf{User Input:}\par
{\ttfamily\small
[User]
[Data]
Protocol: \{PROTOCOL\_LABEL\}
Source: \{SOURCE\}
Cultural Term: \{TERM\}
Cultural Explication (Ground Truth): \{EXPLICATION\}
Model Hypothesis: \{HYPOTHESIS\}

[Task]
1. Ignore whether the translation is elegant.
2. Check ONLY if the semantic meaning matches the Explication.
3. If the hypothesis invents a new entity (Hallucination) or refers to a completely wrong concept, mark INVALID.

[Output format]
Reasoning: [Brief analysis]
Decision: VALID or INVALID
}
\end{tcolorbox}
\caption{The Semantic Validator prompt. It acts as a strict "Validity Gate" ($\mathbb{I}_{valid}$), filtering out hallucinations by anchoring evaluation to the Cultural Explication.}
\label{prompt:validator}
\end{figure*}

\begin{figure*}[t]
\begin{tcolorbox}[
  title={Prompt Template: Stage II-A - Fidelity Critic (Nuance Analysis)},
  colframe=black, colback=white, coltext=black,
  boxrule=0.4pt, arc=1.5pt,
  left=3pt,right=3pt,top=3pt,bottom=3pt,
  boxsep=2pt, toptitle=2pt,bottomtitle=2pt,
  fonttitle=\bfseries\small, fontupper=\small
]
\textbf{System Instruction:}
You are a Translation Critic focusing on "Fidelity" and "Pragmatic Equivalence".
Your job is to score how well the Model Hypothesis captures the intended nuance and force, strictly according to the active Protocol.

REFERENCE USAGE RULE:
\begin{itemize}\setlength\itemsep{0pt}
  \item The [Reference Anchor] is just ONE possible translation.
  \item \textbf{DO NOT} penalize the hypothesis for using different words than the reference.
  \item \textbf{DO} judge based on whether the *meaning* and *effect* are equivalent.
\end{itemize}

DYNAMIC GUIDELINE (Protocol-Dependent):
\{DYNAMIC\_INSTRUCTION\}

SCORING SCALE:
\begin{itemize}\setlength\itemsep{0pt}
  \item \textbf{5 (Perfect):} Captures full nuance, tone, and imagery.
  \item \textbf{3 (Acceptable):} Core meaning present, but significant nuance lost (e.g., too generic).
  \item \textbf{1 (Failure):} Severe mistranslation or complete loss of meaning.
\end{itemize}

\textbf{User Input:}\par
{\ttfamily\small
[User]
[Data]
Protocol: \{PROTOCOL\_LABEL\}
Source: \{SOURCE\}
Reference Anchor: \{REFERENCE\}
Model Hypothesis: \{HYPOTHESIS\}

[Task]
Evaluate Fidelity (1-5).
Focus: Does the hypothesis capture the \{PROTOCOL\_LABEL\} constraints?

[Output format]
Reasoning: [Critique]
Score: [1-5]
}
\end{tcolorbox}
\caption{The Fidelity Critic prompt. It dynamically adjusts criteria (e.g., penalizing literalism in Protocol B vs. generalization in Protocol A) to assess translation nuance.}
\label{prompt:fidelity}
\end{figure*}

\begin{figure*}[t]
\begin{tcolorbox}[
  title={Prompt Template: Stage II-B - Clarity Critic (Communicative Intelligibility)},
  colframe=black, colback=white, coltext=black,
  boxrule=0.4pt, arc=1.5pt,
  left=3pt,right=3pt,top=3pt,bottom=3pt,
  boxsep=2pt, toptitle=2pt,bottomtitle=2pt,
  fonttitle=\bfseries\small, fontupper=\small
]
\textbf{System Instruction:}
You are a Target Audience Evaluator simulating an English native reader who has NO prior knowledge of Chinese culture.
Your goal is to assess "Communicative Intelligibility".

SCORING PHILOSOPHY (Thick Translation):
\begin{itemize}\setlength\itemsep{0pt}
  \item \textbf{Bonus (+):} Reward **Explicitation** (e.g., in-text glosses, brief explanations, appositives like "Kang, a heated brick bed"). Reward **Transparency** (rephrasing for clarity).
  \item \textbf{Penalty (-):} Penalize **Opaque Terms** (Pinyin without context). Penalize **Confusion** (if the reader would ask "What does that mean?").
\end{itemize}

SCORING SCALE:
\begin{itemize}\setlength\itemsep{0pt}
  \item \textbf{5 (Crystal Clear):} Seamlessly bridged. The naive reader fully understands.
  \item \textbf{3 (Gist Only):} Reader gets the general idea but misses the cultural specific.
  \item \textbf{1 (Incomprehensible):} Complete communication breakdown.
\end{itemize}

\textbf{User Input:}\par
{\ttfamily\small
[User]
[Data]
Source: \{SOURCE\}
Model Hypothesis: \{HYPOTHESIS\}

[Task]
Evaluate Clarity (1-5) for a non-Chinese reader.
Did the translator build a bridge for the reader, or leave them confused?

[Output format]
Reasoning: [Analysis]
Score: [1-5]
}
\end{tcolorbox}
\caption{The Communicator prompt. It rewards explicitation strategies consistent with Appiah's "Thick Translation" theory.}
\label{prompt:communicator}
\end{figure*}

\begin{table*}[h]
\centering
\small
\begin{adjustbox}{width=\textwidth,keepaspectratio}
\begin{tabular}{l *{7}{c}}
\toprule
\multirow{2}{*}{\textbf{Benchmark}} 
& \multicolumn{1}{c}{\textbf{Multiple}} 
& \multicolumn{1}{c}{\textbf{Culture-Specific}} 
& \multicolumn{1}{c}{\textbf{Figurative or}} 
& \multicolumn{1}{c}{\textbf{Context-Dependent}} 
& \multicolumn{1}{c}{\textbf{Explicit Error}} 
& \multicolumn{1}{c}{\textbf{Error-Level}} 
& \multicolumn{1}{c}{\textbf{Evaluation Beyond}} \\
& \multicolumn{1}{c}{\textbf{Expression Types}} 
& \multicolumn{1}{c}{\textbf{Meaning}} 
& \multicolumn{1}{c}{\textbf{Non-Literal Meaning}} 
& \multicolumn{1}{c}{\textbf{Interpretation}} 
& \multicolumn{1}{c}{\textbf{Taxonomy}} 
& \multicolumn{1}{c}{\textbf{Analysis}} 
& \multicolumn{1}{c}{\textbf{Surface Metrics}} \\
\midrule
CHENGYU-BENCH 
& \ding{55} & \ding{51} & \ding{51} & \ding{55} & \ding{55} & \ding{55} & \ding{55} \\
IdioTS        
& \ding{55} & \ding{55} & \ding{51} & \ding{51} & \ding{55} & \ding{55} & \ding{55} \\
SlangDIT      
& \ding{55} & \ding{51} & \ding{51} & \ding{51} & \ding{55} & \ding{55} & \ding{51} \\
CAMT          
& \ding{51} & \ding{51} & \ding{55} & \ding{51} & \ding{55} & \ding{55} & \ding{51} \\
IdiomEval     
& \ding{55} & \ding{51} & \ding{51} & $\triangle$ & \ding{51} & \ding{51} & \ding{51} \\
\textbf{CulT-Eval} 
& \ding{51} & \ding{51} & \ding{51} & \ding{51} & \ding{51} & \ding{51} & \ding{51} \\
\bottomrule
\end{tabular}
\end{adjustbox}
\caption{
Comparison of CulT-Eval with existing culture-related translation benchmarks.
 \ding{51}: supported, \ding{55}: not supported, $\triangle$: partially supported.
}
\end{table*}

\subsection{Part Two}

Etiam commodo feugiat nisl pulvinar pellentesque. Etiam auctor sodales
ligula, non varius nibh pulvinar semper. Suspendisse nec lectus non
ipsum convallis congue hendrerit vitae sapien. Donec at laoreet
eros. Vivamus non purus placerat, scelerisque diam eu, cursus
ante. Etiam aliquam tortor auctor efficitur mattis.

\section{Online Resources}

Nam id fermentum dui. Suspendisse sagittis tortor a nulla mollis, in
pulvinar ex pretium. Sed interdum orci quis metus euismod, et sagittis
enim maximus. Vestibulum gravida massa ut felis suscipit
congue. Quisque mattis elit a risus ultrices commodo venenatis eget
dui. Etiam sagittis eleifend elementum.

Nam interdum magna at lectus dignissim, ac dignissim lorem
rhoncus. Maecenas eu arcu ac neque placerat aliquam. Nunc pulvinar
massa et mattis lacinia.

\end{document}